\documentclass{article}

\usepackage{arxiv}
\usepackage[utf8]{inputenc}
\usepackage[T1]{fontenc}
\usepackage{hyperref}
\usepackage{url}
\usepackage{booktabs}      
\usepackage{amsfonts}
\usepackage{nicefrac}
\usepackage{microtype}
\usepackage{lipsum}
\usepackage{graphicx}
\usepackage{amsmath}
\usepackage{algorithm}
\usepackage{algorithmic}
\usepackage{multirow}    

\graphicspath{ {./images/} }

\title{Mortgage Language Model: Domain-Adaptive Pretraining with Residual Instruction, Alignment Tuning, and Task-Specific Routing}

\author{
  \large \textbf{Manish Jain\textsuperscript{*}, Satheesh Kumar Ponnambalam\textsuperscript{*}, Salman Faroz\textsuperscript{*}} \\
  \large \textbf{Chandrakanth Lns\textsuperscript{\dag}, Vinay Sharma\textsuperscript{\ddag}} \\[3pt]
  \normalsize Firstsource \\[3pt]
  \normalsize \texttt{\{Manish.Jain1, Satheeshkumar.Ponnambalam, Salman.Thamimul,} \\
  \normalsize \texttt{Chandrakanth.Lns, Vinay.Sharma4\}@firstsource.com}
}

\begin{document}

\maketitle

\begingroup
\renewcommand\thefootnote{}
\footnotetext{\textsuperscript{*}Equal contribution}
\footnotetext{\textsuperscript{\dag}Supporting contribution}
\footnotetext{\textsuperscript{\ddag}Minor contribution}
\endgroup

\pagestyle{plain}
\begin{abstract}
Large Language Models (LLMs) demonstrate exceptional capabilities across general domains, yet their application to specialized sectors such as mortgage finance requires domain-specific knowledge augmentation while preserving instruction-following fidelity. We present Mortgage Language Model, a novel domain-specific large language model that addresses this dual challenge. It is developed using a dual-track specialization framework from a single base model (Meta-LLaMA-3.1-8B). We opted for this dual-expert approach as a single multi-task model suffers from performance trade-offs, where optimizing for structured tasks (like classification via SFT) degrades conversational fidelity (achieved via DPO). Our dual-track method solves this by creating two specialists, allowing each to be optimally trained for its distinct capability. Our approach applies the instruction residual technique to restore instruction-following capabilities post-domain adaptation without supervised fine-tuning. We contribute: (1) application of this residual technique to the highly specialized mortgage finance domain; (2) a dual-expert architecture combining a conversational Q\&A model and a structured task model for classification and summarization; and (3) an intelligent task routing mechanism using few-shot classification performed by one of the expert models itself. We validate our approach on domain-specific benchmarks, where our final model (MLM v2) significantly outperforms the base LLaMA 3.1 8B Instruct model, achieving an LLM-as-a-Judge summarization score of 4.58 (vs. 3.99), a Q\&A score of 4.09 (vs. 4.0), and a classification score of 2.6 (vs. 1.2). On semantic similarity, our model achieved a BERTScore of 0.77 for summarization (vs. 0.74), 0.68 for Q\&A (vs. 0.58), and 0.75 for classification (vs. 0.73), substantially outperforming baseline approaches. Our framework demonstrates that specialized domain knowledge and general instruction-following abilities can be effectively balanced through the instruction residual methodology in highly technical domains.
\end{abstract}

\section{Introduction}

The widespread adoption of Large Language Models has revolutionized natural language processing across diverse applications. However, applying general-purpose LLMs to specialized domains presents a fundamental challenge: while continued pretraining (CPT) on domain-specific corpora enhances domain knowledge, it frequently leads to catastrophic forgetting \cite{kirkpatrick2017overcoming} of general capabilities, particularly instruction-following behavior acquired during initial instruction-tuning phases.

The mortgage finance domain exemplifies this challenge. Mortgage documents contain specialized terminology, complex regulatory language, and domain-specific conventions that general-purpose models struggle to interpret. Existing approaches to domain adaptation rely primarily on either: (a) expensive supervised fine-tuning (SFT) that risks degrading general capabilities, or (b) retrieval-augmented generation (RAG) that adds latency and complexity. We propose an alternative framework that leverages algebraic manipulation of model weights to preserve instruction-following while enhancing domain knowledge.

We adopt the instruction residual technique introduced by Jindal et al. \cite{jindal2024balancing}, which we refer to as Instruction Residual (IR). This technique exploits the observation that instruction-following capabilities learned during instruction-tuning can be represented as an additive residual in weight space. By extracting this residual from instruction-tuned models and algebraically adding it to domain-adapted models, we restore instruction-following without costly retraining. While Jindal et al. introduced this technique for general domain adaptation, we demonstrate its effectiveness in the specialized mortgage finance domain.

Additionally, we develop a dual-expert architecture where different training strategies optimize for different task types, and implement an intelligent routing system that uses one of the expert models itself as a few-shot classifier to direct queries to the appropriate specialist.

\textbf{Our key contributions are:}
\begin{itemize}
    \item Development of a large-scale, proprietary mortgage document corpus for domain-specific LLM training, encompassing a diverse range of document categories as detailed in Table \ref{tab:corpus_categories_percent}.
    \item Application of instruction residuals (IR) \cite{jindal2024balancing} to mortgage domain adaptation, demonstrating effectiveness in specialized financial domains.
    \item A dual-expert architecture combining conversational and structured task specialists, achieving superior performance across complementary task types.
    \item An intelligent self-routing system where one expert model performs few-shot task classification to route queries efficiently.
\end{itemize}

\begin{table}[h]
\centering
\caption{Distribution of Total Pages Across Document Categories}
\label{tab:corpus_categories_percent}
\small
\begin{tabular}{@{}llll@{}}
\toprule
\multicolumn{4}{c}{\textbf{Document Categories and Percentage of Total Pages}} \\
\midrule
UW (19.6\%) & FNMA (15.9\%) & Case Study (14.5\%) & FHLMC (11.6\%) \\
FHA (8.5\%) & GNMA (7.4\%) & NQM Guidelines (4.8\%) & Servicing (4.7\%) \\
Compliance (3.2\%) & USDA Guidelines (2.9\%) & VA (2.7\%) & VA Guidelines (2.7\%) \\
Compliance Mortgage (0.9\%) & Preprocess Training (0.3\%) & Title (0.2\%) & UW Guidelines (0.1\%) \\
\bottomrule
\end{tabular}
\end{table}

\section{Related Work}

\textbf{Domain Adaptation:} Continued pretraining approaches \cite{gururangan2020dont} improve domain-specific performance but often suffer from catastrophic forgetting. Parameter-efficient fine-tuning methods like LoRA \cite{hu2021LoRA} reduce trainable parameters but still require substantial supervised data and computation.

\textbf{Model Merging:} Recent work on weight interpolation \cite{wortsman2022model} and task arithmetic \cite{ilharco2022editing} suggests that different training objectives create compositional structures in weight space. Jindal et al. \cite{jindal2024balancing} introduced the instruction residual technique, demonstrating that instruction-following capabilities can be extracted as a residual and transferred to domain-adapted models. They show that instruction fine-tuning and continued pre-training operate on approximately orthogonal subspaces, enabling linear combination without catastrophic interference. Our work applies and validates their approach in the mortgage finance domain, extending it to a dual-expert architecture with intelligent routing.

\textbf{Mixture-of-Experts:} Task routing and MoE architectures \cite{shazeer2017outrageously} demonstrate benefits for multi-task learning. However, existing approaches typically train expert models jointly or use dense routing mechanisms. Our lightweight classification-based routing differs by using few-shot task classification from the same model family, reducing overhead while maintaining flexibility.

\section{Methodology}

\subsection{Problem Formulation}

Let $W_{\text{base}} \in \mathbb{R}^{d}$ denote the parameter vector of a base instruction-tuned LLM with $d$ total parameters. We seek to obtain domain-specialized models $W_{\text{Q\&A}}$ and $W_{\text{struct}}$ that minimize distinct, composite loss functions.

The objective for the Q\&A model (Track 1), which uses CPT, Instruction Residual, and DPO, is:
\begin{equation}
\begin{aligned}
W_{\text{Q\&A}} &= \arg\min_W \mathcal{L}_{\text{domain}}(W) + \mathcal{L}_{\textbf{pref}}(W) \\
\text{subject to} \quad & \text{Instruct}(W) \approx \text{Instruct}(W_{\text{base}})
\end{aligned}
\end{equation}

The objective for the structured task model (Track 2), which uses CPT, SFT, and DPO, is:
\begin{equation}
\begin{aligned}
W_{\text{struct}} &= \arg\min_W \mathcal{L}_{\text{domain}}(W) + \mathcal{L}_{\textbf{sft}}(W) + \mathcal{L}_{\textbf{pref}}(W) \\
\text{where} \quad & \mathcal{L}_{\textbf{sft}} = \mathcal{L}_{\text{clf}} + \mathcal{L}_{\text{sum}}
\end{aligned}
\end{equation}

where $\mathcal{L}_{\text{domain}}$ is the domain-specific pretraining loss (from CPT), $\mathcal{L}_{\text{sft}}$ is the supervised fine-tuning loss for structured tasks, $\mathcal{L}_{\text{pref}}$ is the preference alignment loss (from DPO), $\mathcal{L}_{\text{clf}}$ is the classification loss, $\mathcal{L}_{\text{sum}}$ is the summarization loss, and $\text{Instruct}(W)$ measures the instruction-following capability restored by Instruction Residual.

\subsection{Model Selection Rationale}

Our choice of Meta-LLaMA-3.1-8B as the base model was guided by a thorough evaluation based on criteria crucial for our domain, summarized in Table \ref{tab:criteria}.

\begin{table}[h]
\centering
\caption{Model Selection Evaluation Criteria}
\label{tab:criteria}
\begin{tabular}{@{}lll@{}}
\toprule
\textbf{Criteria} & \textbf{Importance} & \textbf{Rationale} \\
\midrule
Context Length & High & Mortgage documents often exceed 4k tokens \\
Performance Benchmarks & High & General capability on standard NLP tasks \\
Resource Requirements & Medium & Model size impacts deployment flexibility \\
Open-Source License & High & Commercial usage rights required \\
\bottomrule
\end{tabular}
\end{table}

We compared several leading open-source models (Table \ref{tab:model_comparison}). The LLaMA 3.1 8B model provided the best balance of high performance on general benchmarks (MMLU, ARC), a very large context window (128k tokens), and a manageable size for fine-tuning and deployment. Models with smaller context windows (e.g., SOLAR, MPT, Falcon) were unsuitable for processing lengthy mortgage documents.

\begin{table}[h]
\centering
\caption{Comparison of Leading Open-Source Models Evaluated}
\label{tab:model_comparison}
\small
\begin{tabular}{@{}lcccccll@{}}
\toprule
\textbf{Model} & \textbf{Context Length} & \textbf{MMLU Score} & \textbf{ARC-Challenge} & \textbf{SQuAD} & \textbf{QuAC (F1)} & \textbf{DROP (F1)} & \textbf{AGIEval English} \\
\midrule
\textbf{LLaMA 3.1 8B} & \textbf{128k} & \textbf{66.7\%} & \textbf{79.7\%} & \textbf{77.0\%} & \textbf{44.9\%} & \textbf{59.5\%} & \textbf{47.8\%} \\
LLaMA 3.2 3B & 128k & 58.0\% & 69.1\% & 67.7\% & 42.9\% & 45.2\% & 39.2\% \\
LLaMA 3.2 1B & 128k & 32.2\% & 32.8\% & 49.2\% & 37.9\% & 28.0\% & 23.3\% \\
SOLAR 10.7B & 4096 & - & - & - & - & - & - \\
OLMO 7B & 4097 & - & - & - & - & - & - \\
MPT 7B & 2048 & - & - & - & - & - & - \\
Falcon 7B & 2048 & - & - & - & - & - & - \\
Cerebras GPT & 2048 & - & - & - & - & - & - \\
\bottomrule
\multicolumn{8}{@{}p{14cm}@{}}{\textit{Note:} Models like SOLAR, OLMO, etc., were disqualified early due to limited context windows, a critical failure point for mortgage document processing. LLaMA 3.1 8B demonstrates the best overall performance among viable candidates.}
\end{tabular}
\end{table}

To further validate our model choice, we evaluated the summarization and semantic capabilities of the LLaMA model family \textbf{using our private, domain-specific evaluation dataset}. The results, detailed in Table \ref{tab:rouge_scores}, confirmed the LLaMA 3.1 8B variant as the most capable, achieving superior ROUGE and BERT\_F1 scores.

\begin{table}[h]
\centering
\caption{Additional Summarization and Semantic Benchmarks for LLaMA Models (on private Data)}
\label{tab:rouge_scores}
\begin{tabular}{@{}lcccc@{}}
\toprule
\textbf{Model} & \textbf{BERT\_F1} & \textbf{ROUGE1} & \textbf{ROUGE2} & \textbf{ROUGEL} \\
\midrule
LLaMA 3.2\_1B & 0.818122 & 0.09555 & 0.021129 & 0.072958 \\
LLaMA 3.2\_3B & 0.839784 & 0.175922 & 0.047071 & 0.139416 \\
\textbf{LLaMA 3.1\_8B} & \textbf{0.843405} & \textbf{0.190669} & \textbf{0.050543} & \textbf{0.149259} \\
\bottomrule
\end{tabular}
\end{table}

\subsection{Data Corpus Construction}

We assembled a proprietary corpus from the mortgage finance domain, consisting of thousands of documents and totaling tens of millions of tokens, with key categories detailed in Table \ref{tab:corpus_categories_percent}.

\subsubsection{Preprocessing and PII Redaction Pipeline}

The raw corpus consisted of heterogeneous file types, including PDF, DOCX, PPTX, image-based documents (e.g., JPG, PNG), and plain text. We developed a robust extraction pipeline to handle this variety, using OCR where necessary, and standardized the output.

For efficient processing and storage, all extracted text was serialized into the Apache Parquet format. GPU-accelerated preprocessing using the NVIDIA cuDF\cite{cudf_github} library significantly reduced processing overhead compared to CPU-based alternatives, as detailed below.

The data was processed through the following critical steps:

\begin{enumerate}
    \item \textbf{Duplicate Removal:} Prior to any cleaning, we performed exact deduplication at the document level. We computed a hash for each document's raw content using the \texttt{blake3} hashing algorithm. Documents with duplicate hashes were discarded, retaining only a single unique copy.

    \item \textbf{Text Normalization and Cleaning:} We implemented a two-stage, hybrid CPU/GPU processing pipeline to clean and normalize the text:
    \begin{itemize}
        \item \textbf{Stage 1 (CPU-based \texttt{ftfy}):} Each document's text was passed through a comprehensive Unicode cleaning function. This process, leveraging the \texttt{ftfy} library, fixed encoding errors, unescaped HTML, uncurled quotes, fixed line breaks, removed control characters, and standardized the text to NFC (Normalization Form C).
        \item \textbf{Stage 2 (GPU-based \texttt{cudf}):} The batch of cleaned text, loaded in a \texttt{cudf} (GPU DataFrame), then underwent high-throughput, GPU-accelerated regex operations. This included the removal of all URLs and the standardization of domain-specific formats (e.g., dates, currency).
    \end{itemize}

    \item \textbf{PII Redaction:} We implemented a high-fidelity PII redaction module. This involved identifying over 20 distinct entities (e.g., names, addresses, social security numbers) and replacing them using the \texttt{faker} library. A key feature was \textit{consistent entity replacement} within a given document chunk; for example, every instance of "John Doe" would be replaced with the same generated male name (e.g., "Michael Smith") throughout that chunk to maintain semantic coherence.

    \item \textbf{Document Segmentation (Chunking):} The cleaned and anonymized documents were decomposed into coherent semantic units using domain-aware chunking strategies to respect document boundaries and logical sections.

\end{enumerate}

\subsection{Initial Baseline - Mortgage Language Model (MLM v1)}

Before developing the dual-track specialization, we established an initial baseline, designated \textbf{MLM v1}. This model was created by applying LoRA-based Supervised Fine-Tuning (SFT) directly to the base Meta-Llama-3.1-8B-Instruct model. The SFT dataset consisted of approximately 29,000 question-answer pairs specific to the mortgage domain.

As shown in Table \ref{tab:mlm_v1_results}, this initial fine-tuning yielded moderate improvements in precision, recall, and F1 score over the base instruction-tuned model.

\begin{table}[h]
\centering
\caption{MLM v1 (LoRA Fine-Tune) vs. Base Model Performance}
\label{tab:mlm_v1_results}
\begin{tabular}{@{}lcc@{}}
\toprule
\textbf{Metric} & \textbf{Base Model (LLaMA 3.1 8B Instruct)} & \textbf{Fine-Tuned (MLM v1)} \\
\midrule
Precision (Relevance) & 51.39 & 53.80 \\
Recall (Completeness) & 68.54 & 70.63 \\
F1 Score & 58.49 & 60.87 \\
\bottomrule
\end{tabular}
\end{table}

However, we observed that while SFT improved performance on in-domain Q\&A, it did not sufficiently imbue the model with deep domain-specific terminology and context from the raw corpus. This led to hallucinations on complex, document-grounded queries. To improve the model's fundamental domain knowledge, we moved to a Continued Pre-training (CPT) approach, which necessitated the more advanced dual-track methodology to subsequently restore instruction-following capabilities.

\subsection{Dual-Track Model Specialization - Mortgage Language Model (MLM v2)}

Both expert models begin with the same foundation: Meta-LLaMA-3.1-8B base model undergoes Continued Pre-training (CPT) on our mortgage corpus. However, the subsequent specialization paths diverge based on task requirements.

\subsubsection{Track 1: Conversational Q\&A Specialist via Instruction Residual}

\begin{figure}[H]
\centering
\includegraphics[width=\textwidth]{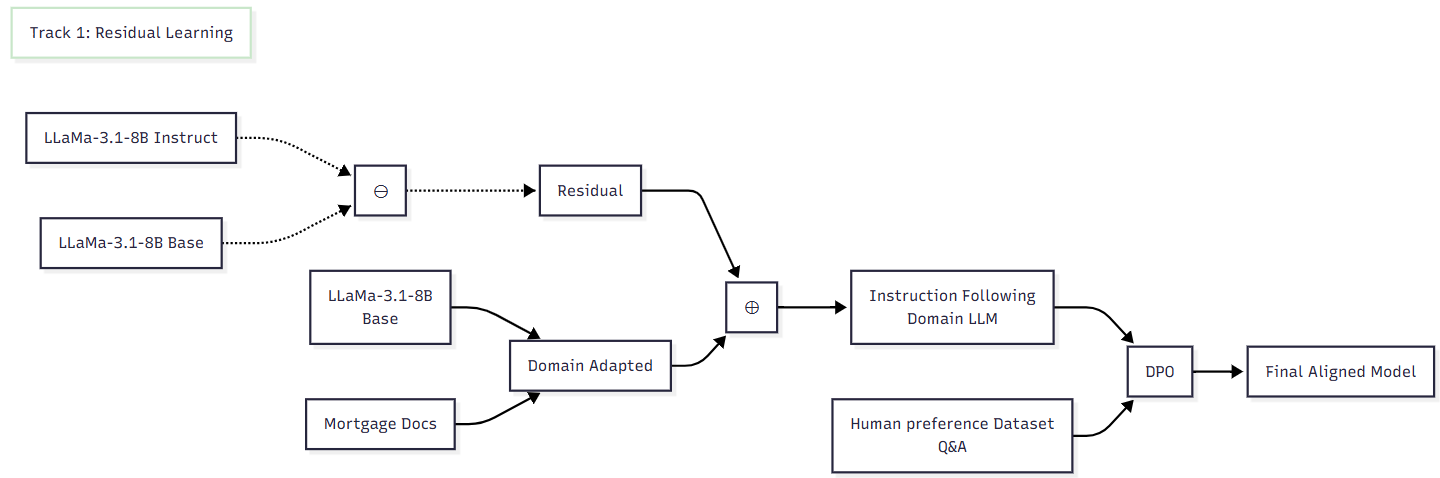}
\caption{Architectural flow for Track 1 (Residual Learning).}
\label{fig:track1_flow}
\end{figure}

\textbf{Step 1 - Continued Pre-training:}
Our CPT journey begins with the foundational Meta-LLaMA-3.1-8Bs base model ($W_{\text{base}}$). This step is critical, as our initial baseline (MLM v1) revealed that supervised fine-tuning alone was insufficient to imbue the model with deep domain knowledge. To address this, we perform CPT on our proprietary domain corpus ($\mathcal{D}_{\text{mortgage}}$), which consists of millions of specialized tokens. The objective is to shift the model's internal representations from general web text to the specific terminology, regulations, and contexts of the mortgage finance domain. This phase produces a domain-specialized, non-instruction-following model, $W_{\text{cpt}}^{(1)}$, optimized via the standard causal language modeling loss. The specific hyperparameters for this CPT stage are detailed in Table \ref{tab:cpt_sft_params}.

\begin{equation}
W_{\text{cpt}}^{(1)} = \arg\min_W \sum_{x \in \mathcal{D}_{\text{mortgage}}} \mathcal{L}_{\text{lm}}(x; W)
\end{equation}

where $\mathcal{L}_{\text{lm}}$ is the causal language modeling loss (next-token prediction).

\textbf{Step 2 - Instruction Residual (IR):}
We adopt the instruction residual technique introduced by Jindal et al. \cite{jindal2024balancing}, which we term Instruction Residual (IR). This technique demonstrates that instruction-following capabilities can be represented as an additive residual in weight space and transferred to domain-adapted models without requiring Instruction tuning.

Following their methodology, we compute the instruction residual from two reference models in the same architecture family - $W_{\text{LLaMA-base}}$ (pretrained only) and $W_{\text{LLaMA-inst}}$ (instruction-tuned):

\begin{equation}
\Delta_{\text{inst}} = W_{\text{LLaMA-inst}} - W_{\text{LLaMA-base}}
\end{equation}

This residual $\Delta_{\text{inst}}$ captures the weight changes necessary to acquire instruction-following behavior. Following the direct addition method proposed by Jindal et al. \cite{jindal2024balancing}, we apply this residual to our domain-adapted model:

\begin{equation}
W_{\text{IR}} = W_{\text{cpt}}^{(1)} + \Delta_{\text{inst}}
\end{equation}

\textbf{Theoretical Foundation from Jindal et al.:}
The Instruction Residual approach is grounded in the observation that continued pre-training and instruction fine-tuning operate on approximately orthogonal subspaces in the model's weight space \cite{jindal2024balancing}. This enables:
\begin{enumerate}
    \item \textit{Transferability}: The instruction residual $\Delta_{\text{inst}}$ generalizes across different training distributions within the same architecture family, as it encodes task-agnostic instruction-following behavior.
    \item \textit{Compositionality}: Domain knowledge acquired through CPT and instruction-following capabilities can be linearly combined without significant catastrophic interference.
\end{enumerate}

Jindal et al. demonstrate that this approach successfully balances domain expertise with general instruction-following, which we validate in the mortgage finance domain.

\textbf{Step 3 - Direct Preference Optimization (DPO)\cite{rafailov2023dpo} with LoRA:}
Finally, we apply DPO using Low-Rank Adaptation (LoRA) to align outputs with human preferences for conversational quality. This provides a computationally efficient alignment, as detailed in Section \ref{sec:dpo_details}, by freezing the $W_{\text{IR}}$ weights and training only low-rank adapters. The hyperparameters for this DPO alignment are specified in Table \ref{tab:dpo_params}. The objective is:

\begin{equation}
W_{\text{Q\&A}} = \arg\min_W \mathbb{E}_{(x, y^+, y^-) \sim \mathcal{D}_{\text{pref}}} \left[ -\log \sigma\left( \beta \log \frac{\pi_W(y^+ | x)}{\pi_{\text{ref}}(y^+ | x)} - \beta \log \frac{\pi_W(y^- | x)}{\pi_{\text{ref}}(y^- | x)} \right) \right]
\end{equation}

where $y^+$ and $y^-$ are preferred and dispreferred completions, $\beta$ is the inverse temperature parameter, $\sigma$ is the sigmoid function, $\pi_W$ is the policy being optimized (with LoRA adapters), and $\pi_{\text{ref}} = W_{\text{IR}}$ serves as the frozen reference policy.

The complete training pipeline for Model 1 (MLM-Q\&A) is:

\begin{equation}
\boxed{W_{\text{Q\&A}} = \text{DPO-LoRA}\left(W_{\text{cpt}}^{(1)} + \Delta_{\text{inst}}, \mathcal{D}_{\text{pref}}\right)}
\end{equation}

\subsubsection{Track 2: Structured Task Specialist via SFT}

\begin{figure}[H]
\centering
\includegraphics[width=\textwidth]{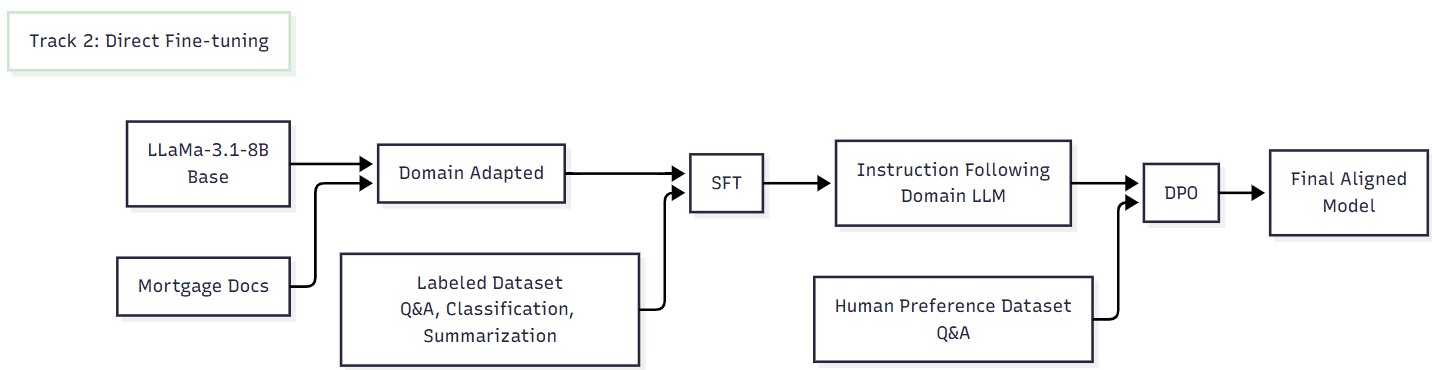}
\caption{Architectural flow for Track 2 (Direct Fine-tuning).}
\label{fig:track2_flow}
\end{figure}

\textbf{Step 1 - Continued Pre-training:}
Similar to Track 1, the foundation for our structured task specialist is a model deeply embedded with domain knowledge. We perform an identical CPT process, starting from the same $W_{\text{base}}$ and training on the full $\mathcal{D}_{\text{mortgage}}$ corpus. This ensures that the model, prior to any task-specific fine-tuning, has already learned the nuances of mortgage terminology. This process yields $W_{\text{cpt}}^{(2)}$, a domain-adapted base model. As noted in Track 1, the specific hyperparameters for this CPT stage are detailed in Table \ref{tab:cpt_sft_params}.

\textbf{Step 2 - Supervised Fine-Tuning with LoRA:}
For tasks requiring specific output formats (classification, summarization), we create a multi-task supervised dataset:

\begin{equation}
\mathcal{D}_{\text{struct}} = \mathcal{D}_{\text{clf}} \cup \mathcal{D}_{\text{sum}} \cup \mathcal{D}_{\text{Q\&A-sup}}
\end{equation}

where $\mathcal{D}_{\text{clf}}$ contains classification examples, $\mathcal{D}_{\text{sum}}$ contains summarization examples, and $\mathcal{D}_{\text{Q\&A-sup}}$ contains structured Q\&A examples.

We employ Low-Rank Adaptation (LoRA) to efficiently fine-tune the model. LoRA freezes the pretrained weights $W_{\text{cpt}}^{(2)}$ and introduces trainable low-rank decomposition matrices:

\begin{equation}
W_{\text{sft}} = W_{\text{cpt}}^{(2)} + \Delta W = W_{\text{cpt}}^{(2)} + BA
\end{equation}

where $B \in \mathbb{R}^{d \times r}$ and $A \in \mathbb{R}^{r \times k}$ are low-rank matrices with rank $r \ll \min(d, k)$. The optimization objective is:

\begin{equation}
\min_{A,B} \sum_{(x,y) \in \mathcal{D}_{\text{struct}}} \mathcal{L}_{\text{task}}(y, f_{W_{\text{cpt}}^{(2)} + BA}(x))
\end{equation}

LoRA reduces trainable parameters from $O(d \times k)$ to $O(r(d+k))$, enabling efficient fine-tuning while preserving domain knowledge acquired during CPT.

Table \ref{tab:cpt_sft_params} outlines the hyperparameters for the initial domain adaptation (CPT) and supervised fine-tuning (SFT) stages. The **Min/Max Token Length** parameters define the bounds for the input data sequences, ensuring the model processes chunks of a consistent and meaningful size (e.g., 419-2741 tokens for Track 1 CPT). We used a high LoRA rank (256) for the CPT stages to capture a wide range of domain knowledge, targeting nearly all module types. For SFT, a much smaller rank (16) was used to specialize the model on specific tasks without overwriting the CPT knowledge.

\begin{table}[H]
\centering
\caption{Hyperparameters for CPT and SFT Stages}
\label{tab:cpt_sft_params}
\small
\begin{tabular}{@{}llll@{}}
\toprule
\textbf{Parameter} & \textbf{Track 1 — CPT} & \textbf{Track 2 — CPT} & \textbf{Track 2 — SFT} \\
\midrule
LoRA Rank (r) & 256 & 256 & 16 \\
LoRA Alpha & 256 & 256 & 32 \\
LoRA Dropout & 0.15 & 0.15 & 0.2 \\
Target Modules & \begin{tabular}[t]{@{}l@{}}q\_proj, k\_proj, v\_proj, \\ o\_proj, gate\_proj, \\ up\_proj, down\_proj, \\ embed\_tokens, lm\_head\end{tabular} & \begin{tabular}[t]{@{}l@{}}q\_proj, k\_proj, v\_proj, \\ o\_proj, gate\_proj, \\ up\_proj, down\_proj, \\ embed\_tokens, lm\_head\end{tabular} & \begin{tabular}[t]{@{}l@{}}q\_proj, v\_proj\end{tabular} \\
Learning Rate & 0.0002 & 0.0002 & 0.001 \\
Epochs & 2 & 1 & 3 \\
Training Time & 3 hours & 1h 18m & 8h 55m \\
Min Token Length & 419 & 419 & 31 \\
Max Token Length & 2741 & 2086 & 2793 \\
\bottomrule
\end{tabular}
\end{table}

\textbf{Step 3 - Direct Preference Optimization (DPO) with LoRA:}
Similar to Track 1, we apply DPO using LoRA for preference-based refinement, following the methodology described in Section \ref{sec:dpo_details}. The hyperparameters for this DPO alignment are specified in Table \ref{tab:dpo_params}.

\begin{equation}
W_{\text{struct}} = \text{DPO-LoRA}(W_{\text{sft}}, \mathcal{D}_{\text{pref-struct}})
\end{equation}

The complete training pipeline for Model 2 (MLM-Clf/Sum) is:

\begin{equation}
\boxed{W_{\text{struct}} = \text{DPO-LoRA}\left(W_{\text{cpt}}^{(2)} + BA, \mathcal{D}_{\text{pref-struct}}\right)}
\end{equation}

\subsubsection{Alignment with Direct Preference Optimization (DPO)}
\label{sec:dpo_details}

Both specialization tracks conclude with an alignment phase using Direct Preference Optimization (DPO)\cite{rafailov2023dpo}. DPO replaces the complex reward-model training and RLHF fine-tuning with a closed-form objective that directly optimizes the policy on preference pairs. This approach simplifies alignment while maintaining competitive quality and stability.

\textbf{Dataset Curation Pipeline:}
The foundation of our DPO training is a robust dataset of human preferences derived from various internal sources. This dataset was constructed by creating preference pairs, each consisting of a "chosen" (preferred) and a "rejected" (dispreferred) option from model-generated responses. To ensure data quality, these pairs were filtered based on the rating deltas provided by Subject Matter Experts (SMEs). The final dataset was split into training (85\%) and evaluation (15\%) sets, using a stratified split based on data category to preserve distributions.

\textbf{Model Architecture and Training Setup:}
To make the DPO fine-tuning process efficient, we utilized Low-Rank Adaptation (LoRA). This approach introduces a small number of trainable parameters ($O(r(d+k))$) in the form of low-rank matrices, freezing the vast majority of the foundation model's weights. For this project, the foundation models were the domain-finetuned outputs from Track 1 ($W_{\text{IR}}$) and Track 2 ($W_{\text{sft}}$). We specifically targeted the attention mechanism's query, key, value, and output projections (q\_proj, k\_proj, v\_proj, o\_proj). This allows for significant updates to the model's behavior while minimizing computational cost. A systematic hyperparameter sweep was conducted to identify the optimal training configuration, evidenced by a consistent reduction in training and evaluation loss.

The final DPO hyperparameters for both tracks are detailed in Table \ref{tab:dpo_params}. The **Cutoff length** (2048) specifies the maximum sequence length the model will process for a single DPO training example; any input longer than this is truncated. The **Min/Max context lengths** (161-4991) refer to the statistical range of sequence lengths found in the raw preference dataset, prior to truncation. Key parameters include a low LoRA rank (8) for gentle alignment, a $\beta$ value of 0.2, and distinct learning rates (2.5e-05 for Q\&A, 3e-05 for Struct) to optimize for each track's specific needs.

\begin{table}[H]
\centering
\caption{Hyperparameters for DPO Alignment Stages}
\label{tab:dpo_params}
\small
\begin{tabular}{@{}lll@{}}
\toprule
\textbf{Parameter} & \textbf{Track 1 — DPO (Q\&A)} & \textbf{Track 2 — DPO (Struct)} \\
\midrule
LoRA Rank (r) & 8 & 8 \\
LoRA Alpha & 16 & 16 \\
LoRA Dropout & 0.1 & 0.1 \\
Target Modules & q\_proj, v\_proj, k\_proj, o\_proj & q\_proj, v\_proj, k\_proj, o\_proj \\
$\beta$ value for DPO & 0.2 & 0.2 \\
Min / Max context lengths & 161 and 4991 & 161 and 4991 \\
Cutoff lengths & 2048 & 2048 \\
Epochs & $\sim$1 epoch (0.98) & $\sim$1 epoch \\
Steps Completed & 18000 & 18000 \\
Effective Batch Size & 4 & 4 \\
Learning Rate & 2.5e-05 & 3e-05 \\
GPU Hours & 12 hrs & 33 hrs \\
\bottomrule
\end{tabular}
\end{table}

\subsection{Comparative Analysis of Specialization Tracks}

Table \ref{tab:tracks} summarizes the key differences between the two specialization tracks:

\begin{table}[h]
\centering
\small
\begin{tabular}{lll}
\toprule
\textbf{Aspect} & \textbf{Track 1 (Q\&A)} & \textbf{Track 2 (Clf/Sum)} \\
\midrule
Base Training & CPT on $\mathcal{D}_{\text{mortgage}}$ & CPT on $\mathcal{D}_{\text{mortgage}}$ \\
Specialization & Instruction Residual & LoRA-based SFT \\
& + DPO (w/ LoRA) & + DPO (w/ LoRA) \\
Data Required & Unsupervised + preferences & Supervised labels + preferences \\
Parameters Added & \textbf{$O(r(d+k))$} trainable & $O(r(d+k))$ trainable \\
Optimal For & Conversational Q\&A & Classification, Summarization \\
\bottomrule
\end{tabular}
\caption{Comparison of dual-track specialization strategies.}
\label{tab:tracks}
\end{table}

\section{Evaluation and Benchmarking}

\subsection{Overview}
To assess the performance of our customized LLaMA-based models (MLM v2), we conducted a comprehensive evaluation and benchmarking study across three natural language understanding (NLU) tasks — Summarization, Classification, and Question Answering (Q\&A). These tasks were selected to capture generative fluency (summarization), semantic discrimination (classification), and contextual comprehension and reasoning (Q\&A).

The evaluation was performed against four baseline models:
\begin{itemize}
    \item \textbf{AdaptLLM:} finance-LLM
    \item \textbf{Mistral:} Mistral-7B-Mortgage-Loans
    \item \textbf{LLaMA 3:} LLaMA 3.1 8B INSTRUCT (Base Model)
    \item \textbf{MLM v1:} LLaMA 3.1 8B INSTRUCT + SFT with LoRA
\end{itemize}

MLM v2 (based on LLaMA 3.1 8B base) represents the combined output of our dual-track system:
\begin{itemize}
    \item \textbf{MLM v2 Track 1:} CPT + RESIDUAL + DPO (Used for Q\&A)
    \item \textbf{MLM v2 Track 2:} CPT + SFT + DPO (Used for Summarization \& Classification)
\end{itemize}

\subsection{Evaluation Methodology}
Our evaluation framework involves BERTScore, LLM-based comparative judgment, Domain Specific Term Evaluation, and human expert (SME) evaluation to ensure a balanced assessment.A Higher evaluation score corresponds to better model performance. All detailed charts and specific results for these evaluations are provided in Appendix A.

\subsection{Results Summary}
Our MLM v2 model consistently outperformed all baseline models across our comprehensive evaluation framework.
\begin{itemize}
    \item \textbf{Semantic Similarity:} MLM v2 demonstrated the highest semantic fidelity, achieving the top scores in the BERTScore evaluation for Q\&A, Summarization, and Classification (see Appendix, Figure \ref{fig:bert_score}).
    \item \textbf{LLM-as-a-Judge:} In the scalable LLM-as-a-Judge evaluation, our model again achieved the highest ratings, notably 4.58 for summarization and 2.6 for classification (see Appendix, Figure \ref{fig:llm_judge}).
    \item \textbf{Domain-Specific Knowledge:} MLM v2 showed superior domain alignment, scoring highest on all three specialized metrics: Mortgage MCQ Accuracy (64.1\%), GPT-as-a-Judge (3.62), and Mortgage Key Terms BERTScore (0.70) (see Appendix, Figure \ref{fig:mortgage_combined_scores}).
    \item \textbf{Human Expert (SME) Evaluation:} The results were overwhelmingly in favor of our model, with 92.9\% of domain experts preferring the outputs from MLM v2 over all baselines (see Appendix, Figure \ref{fig:sme_eval}).
    \item \textbf{Security:} MLM v2 demonstrated a significantly improved security posture, with substantial gains in Prompt Injection (66.4\%) and Malware Generation (80.7\%) resistance, while maintaining perfect safety (100.0\%) on PII and Toxicity benchmarks (see Appendix, Figure \ref{fig:security_metrics}).
\end{itemize}

\section{Intelligent Task Routing System}

\subsection{Self-Routing Architecture}

To operationalize the dual-expert system, we implement an intelligent routing mechanism. we use MLM-Q\&A (Model 1) itself as the task classifier, eliminating the need for a separate classification model.

\subsection{Few-Shot Task Classification}

Upon receiving a user query $q$, the router constructs a structured classification prompt $p_c(q)$ that includes:
\begin{itemize}
    \item Explicit definitions of three task categories: Q\&A, Classification, Summarization
    \item Few-shot exemplars (in the prompt, not requiring training)
    \item The user query $q$
    \item Instructions to output the category number (1, 2, or 3)
\end{itemize}

The classification is performed by the MLM-Q\&A model itself:

\begin{equation}
c^* = \text{Classify}_{W_{\text{Q\&A}}}(p_c(q))
\end{equation}

where $\text{Classify}_{W_{\text{Q\&A}}}(\cdot)$ denotes the classification function implemented by model $W_{\text{Q\&A}}$ through text generation. The model generates a response indicating the task category, which is then parsed to determine the routing decision.

\subsection{Routing Decision}

Based on the predicted category $c^*$, the routing decision is:

\begin{equation}
\text{Route}(q) = \begin{cases}
W_{\text{Q\&A}} & \text{if } c^* = \text{Q\&A} \\
W_{\text{struct}} & \text{if } c^* \in \{\text{Clf}, \text{Sum}\}
\end{cases}
\end{equation}

The routing algorithm is formalized as:

\begin{algorithm}[h]
\caption{Self-Routing Task Classification}
\begin{algorithmic}[1]
\STATE \textbf{Input:} User query $q$, Expert models $\{W_{\text{Q\&A}}, W_{\text{struct}}\}$
\STATE \textbf{Output:} Selected model $W^*$
\STATE
\STATE Construct classification prompt: $p_c(q) \leftarrow \text{FormatPrompt}(q)$
\STATE Generate classification: $\text{response} \leftarrow W_{\text{Q\&A}}(p_c(q))$
\STATE Parse category: $c^* \leftarrow \text{ParseCategory}(\text{response})$
\STATE
\IF{$c^* \in \{\text{Classification}, \text{Summarization}\}$}
    \STATE $W^* \leftarrow W_{\text{struct}}$
\ELSE
    \STATE $W^* \leftarrow W_{\text{Q\&A}}$
\ENDIF
\STATE
\RETURN $W^*$
\end{algorithmic}
\end{algorithm}

\subsection{Efficiency Analysis}

The self-routing classification step introduces minimal overhead (empirically measured at <100ms) as it reuses the already-deployed $W_{\text{Q\&A}}$ model. The primary efficiency bottleneck in a production environment is typically the inference backend itself.

Traditional inference, implemented through the Hugging Face Transformers API, exhibits a batch-dependent performance ceiling governed by synchronous processing. As batch size increases, throughput scales linearly until GPU saturation, but latency expands proportionally. Our benchmarks of this static-batching approach (based on 100 total requests) show throughput saturating at ~9.5 req/s at a batch size of 128, but this comes at the cost of average latency rising from 6.5s to over 10.4s per request. This reflects the intrinsic limitation of static batching—each generation call must complete before subsequent requests are initiated.

To overcome this, our system uses the vLLM \cite{kwon2023vLLM} runtime, which introduces a dynamic, streaming-oriented inference paradigm. We benchmarked the vLLM \cite{kwon2023vLLM} performance by sending 100 requests to a server configured with an increasing number of concurrent workers, with results detailed in Table \ref{tab:vLLM_benchmarking}.

\begin{table}[H]
\centering
\caption{vLLM Inference Benchmarking (100 Requests)}
\label{tab:vLLM_benchmarking}
\small
\begin{tabular}{@{}cccccc@{}}
\toprule
\textbf{Workers} & \textbf{Throughput} & \textbf{Success} & \textbf{Avg. Latency} & \textbf{P95 Latency} & \textbf{Avg. TTFT} \\
& \textbf{(req/s)} & \textbf{Rate} & \textbf{(s)} & \textbf{(s)} & \textbf{(s)} \\
\midrule
1 & 0.7516 & 1.0 & 1.3289 & 2.6586 & 0.1435 \\
2 & 1.5042 & 1.0 & 1.3239 & 2.7683 & 0.1468 \\
4 & 2.6433 & 1.0 & 1.4615 & 2.8345 & 0.1490 \\
8 & 4.6820 & 1.0 & 1.4536 & 2.9739 & 0.1592 \\
12 & 6.2118 & 1.0 & 1.6042 & 3.0588 & 0.1524 \\
16 & 6.5319 & 1.0 & 1.9126 & 3.9049 & 0.6441 \\
20 & 9.2885 & 1.0 & 1.5640 & 3.1957 & 0.1599 \\
24 & 8.1765 & 1.0 & 1.7763 & 3.3452 & 0.5311 \\
28 & 9.0969 & 1.0 & 1.7277 & 3.2773 & 0.1788 \\
32 & 9.2697 & 1.0 & 1.6665 & 3.2173 & 0.1693 \\
\bottomrule
\end{tabular}
\end{table}

\subsection{Hardware Efficiency Details}

The experiments for DPO fine-tuning and vLLM inference were conducted on the following hardware setup.

\noindent\textbf{GPU Specifications}
\begin{itemize}
    \item \textbf{GPU Model:} NVIDIA A100 80GB PCIe
    \item \textbf{Driver Version:} 570.195.03
    \item \textbf{CUDA Version:} 12.8
    \item \textbf{Total GPU Memory:} 80 GB
    \item \textbf{Total CPU Memory:} 216 GB
\end{itemize}

\begin{table}[H]
\centering
\caption{Input Token Statistics Across Concurrent Requests}
\begin{tabular}{|c|c|c|}
\hline
\textbf{Concurrent Requests} & \textbf{Avg Input Tokens} & \textbf{Total Input Tokens} \\
\hline
4  & 39432 & 157728 \\
8  & 19254 & 154032 \\
16 & 9597  & 153552 \\
32 & 4720  & 151040 \\
\hline
\end{tabular}
\end{table}

\noindent\textbf{Total input token length used for summarization:} 78{,}503

In summary, vLLM \cite{kwon2023vLLM} offers a measurable architectural advancement. Its improvements in latency (achieving a mean latency of $\approx$1.7s) and time-to-first-token (averaging $\approx$0.17s), even under high load with 32 workers, translate into superior efficiency, scalability, and responsiveness. These findings empirically substantiate the transition of the MLM architecture toward stream-optimized inference. Both expert models are served using this vLLM engine—which is specifically designed for high-throughput LLM serving—positioning it as the preferred backend for large-scale, multi-expert LLM deployments in production environments.

\section{Limitations}

While our dual-track framework and the resulting Mortgage Language Model v2 model show significant improvements, we acknowledge several limitations that provide avenues for future work.

\begin{itemize}
    \item \textbf{Corpus Scale and Modality:} Our CPT corpus, while highly specialized, is modest in size (tens of millions of tokens). The model's deep domain knowledge is likely constrained by this data scale, and performance could be further improved by expanding the corpus by an order of magnitude. Furthermore, our current pipeline flattens multi-modal mortgage documents (which include complex tables, layouts, and checkboxes) into raw text, losing valuable structural information that a vision-language model (VLM) could exploit.

    \item \textbf{Inference and Routing Complexity:} The dual-expert architecture, by definition, introduces overhead. It requires serving two distinct model specialists and, more critically, relies on a few-shot classification router. This router, while efficient, is a potential point of failure. A misclassification by the router will direct the query to the sub-optimal expert (e.g., sending a complex Q\&A query to the SFT model), leading to a degraded or incorrectly formatted response.

    \item \textbf{Algorithmic Constraints of DPO\cite{feng2024analyzing}:} Although DPO offers a lightweight and stable alternative to RLHF, it exhibits two fundamental limitations that affect its alignment effectiveness in practice. DPO tends to penalize dispreferred responses more strongly than it rewards preferred ones, causing models to learn avoidance behaviors rather than robustly improving preferred-output quality. Its performance is also highly dependent on the SFT initialization weak SFT separation between preferred and rejected responses leads to slow or unstable preference alignment. This sensitivity limits alignment robustness, particularly when the upstream SFT stage is not sufficiently strong.

    \item \textbf{Dependency on Vendor Models for IR:} The Instruction Residual (IR) technique for Track 1 is highly effective but also dependent on the public availability of *both* the original base model ($W_{\text{LLaMA-base}}$) and its corresponding instruction-tuned version ($W_{\text{LLaMA-inst}}$). This method is not applicable if a model vendor only releases a single version (e.g., only the instruct-tuned model), which limits the generalizability of this specific track.
\end{itemize}

\section{Future Work}

While the Mortgage Language Model framework demonstrates a robust method for balancing domain specialization with instruction-following, several avenues for future research remain.

\begin{itemize}
    \item \textbf{Corpus Expansion and Multi-Modal Integration:}
    The current CPT corpus, while effective at 21M tokens, is a baseline. A key next step is to scale the proprietary corpus by at least an order of magnitude (e.g., to 200M+ tokens) to analyze the scaling laws governing domain knowledge acquisition. Furthermore, mortgage documents are inherently multi-modal, relying on visual layout (tables, checkboxes, signatures). We plan to integrate vision-language models (VLMs) to move from pure text to a richer, layout-aware pretraining process.

    \item \textbf{Evolution to a Sparsely-Gated MoE:}
    The current dual-expert system with a few-shot classifier router is a simple 2-expert MoE. A natural evolution is to expand this into a fully-fledged, sparsely-gated Mixture-of-Experts (MoE) architecture. This would involve training multiple specialist models (e.g., a "Compliance Expert," a "Data Extraction Expert," a "Credit Analysis Expert") from the same CPT base and training a lightweight, learnable router to allocate tokens to the most relevant expert(s) during inference.

    \item \textbf{Advanced Domain-Specific Benchmarking:}
    We plan to develop a more challenging benchmark focused on complex numerical and multi-hop reasoning specific to underwriting. This would include tasks like: (1) Verifying calculations on a Closing Disclosure form; (2) Cross-referencing multiple documents to check for logical consistency (e.g., "Does the appraisal value match the LTV calculation on the loan application?"); and (3) Adversarial testing for numerical perturbations.
\end{itemize}

%Bibliography
\bibliographystyle{unsrt}  
\bibliography{references}

\appendix

\section{Detailed Evaluation Results}
\label{sec:appendix_eval}
This appendix provides the detailed visualizations and descriptions for the evaluation results summarized in the main paper. A comprehensive evaluation of the proposed LLM was performed across three natural language understanding tasks: Question Answering (1,000 samples), Summarization (200 samples), and Text Classification (200 samples).

\subsection{BERTScore Evaluation}
\begin{figure}[h]
\centering
\includegraphics[width=0.9\textwidth]{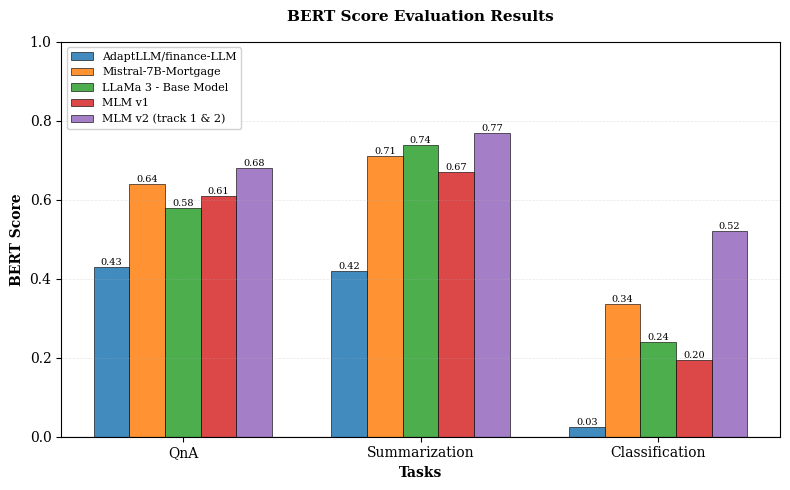}
\caption{BERTScore Evaluation Results (Higher is better)}
\label{fig:bert_score}
\end{figure}

We evaluated BERTScore using ‘microsoft/deberta-xlarge-mnli’ for summarization and Q\&A, measuring semantic similarity between generated and reference responses. Unlike traditional lexical overlap metrics (e.g., ROUGE, BLEU), BERTScore captures contextual and semantic fidelity, aligning closely with human perception of meaning preservation. The comparative results across all models are presented in Figure \ref{fig:bert_score}, where MLM v2 demonstrates the highest scores.

\subsection{LLM-as-a-Judge Evaluation}
\begin{figure}[h]
\centering
\includegraphics[width=0.9\textwidth]{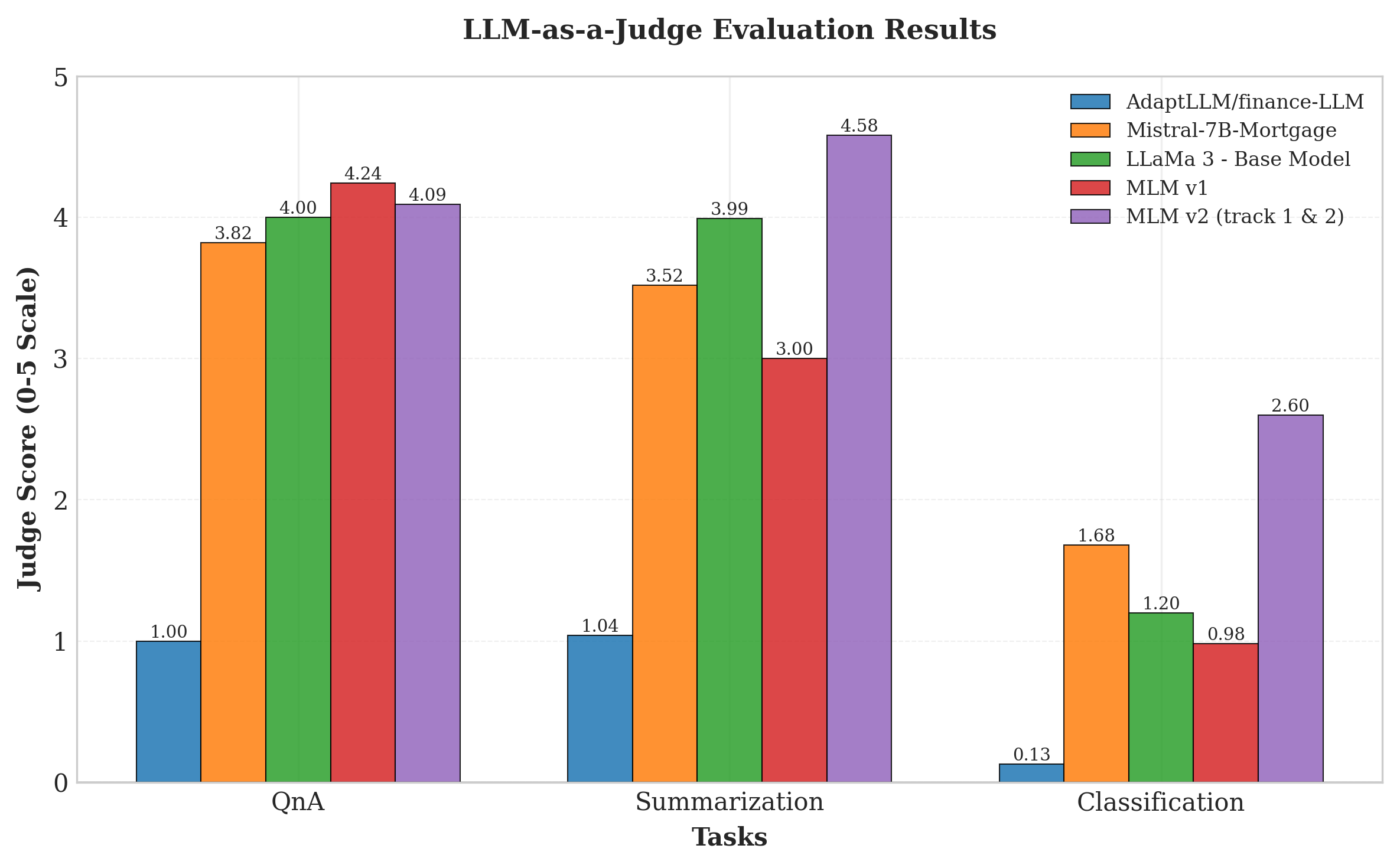}
\caption{LLM-as-a-Judge Evaluation Results (Higher is better)}
\label{fig:llm_judge}
\end{figure}

To complement BERTScore metrics, we used \textbf{frontier model} for an LLM-based comparative evaluation framework. LLM-judge methods provide scalable and relative evaluations, mitigating the subjectivity and variability of human annotators. As shown in Figure \ref{fig:llm_judge}, our MLM v2 model achieved the highest ratings for summarization (4.58) and classification (2.6) among all contenders.

\subsection{SME (Subject Matter Expert) Evaluation}
\begin{figure}[h!]
\centering
\includegraphics[width=0.9\textwidth]{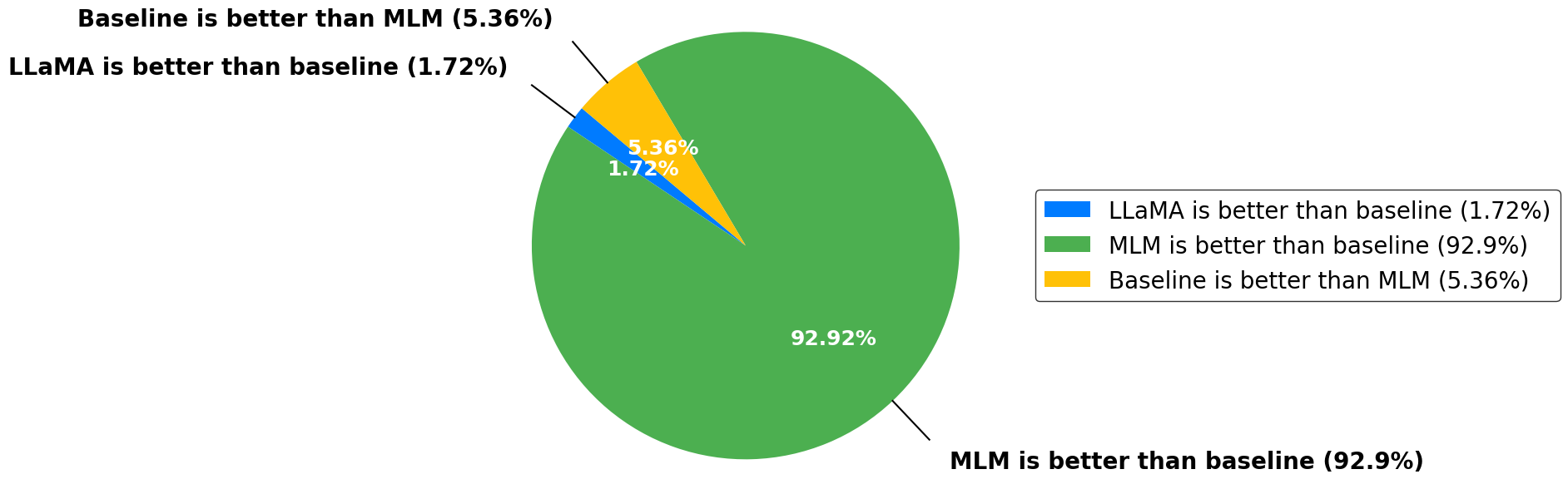}
\caption{SME Model Evaluation Distribution}
\label{fig:sme_eval}
\end{figure}

Human evaluation was conducted by domain SMEs with extensive expertise in financial documentation, policy interpretation, and regulatory compliance. SMEs rated outputs on:
\begin{itemize}
    \item Domain accuracy and factual correctness
    \item Relevance and contextual appropriateness
    \item Readability and professional tone
    \item Faithfulness to source material
\end{itemize}
The results, shown in Figure \ref{fig:sme_eval}, demonstrate an overwhelming preference for the MLM v2 model.

\subsection{Domain Specific Term Evaluation}
We also evaluated all the baseline models and MLM v2 on domain specific mortgage terminology to assess their semantic alignment within the financial domain. This evaluation used three distinct metrics:

\begin{itemize}
    \item \textbf{Mortgage MCQ Accuracy:} Measured the percentage of correctly answered multiple-choice questions, highlighting the improvement of MLM v2 (64.1\%) over its predecessor MLM v1 (47.8\%).
    \item \textbf{GPT-as-a-Judge Mortgage Key:} Scored on a 0–5 scale, this metric evaluated relative answer relevance, where MLM v2 obtained a strong 3.62 score.
    \item \textbf{Mortgage Key Terms – BERTScore:} Scored on a 0–1 scale, this evaluated semantic alignment with reference mortgage key term definitions. MLM v2 achieved the highest score of 0.70.
\end{itemize}

\begin{figure}[h!]
\centering
\includegraphics[width=0.9\textwidth]{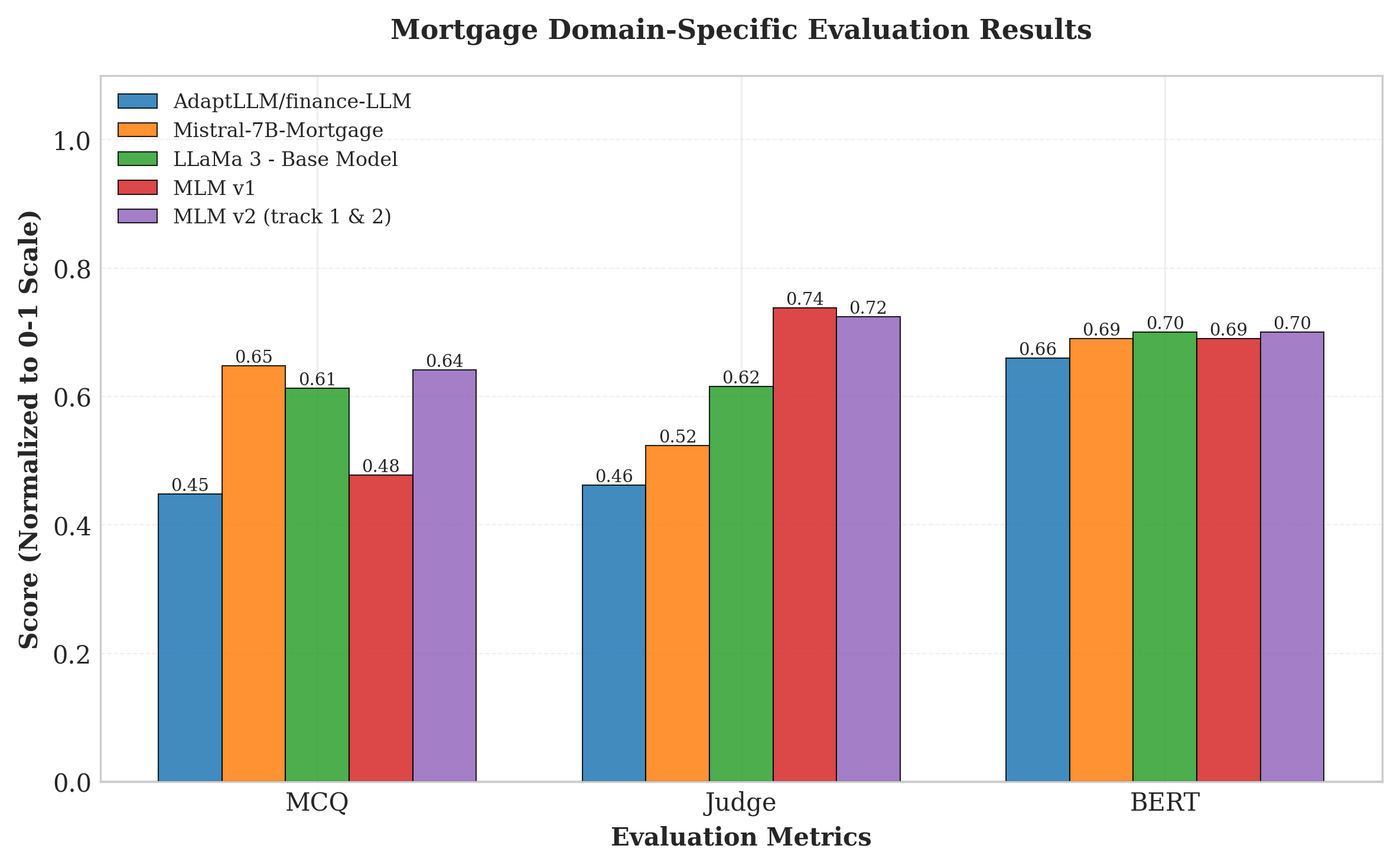}
\caption{Mortgage Evaluation Results (Higher is better)}
\label{fig:mortgage_combined_scores}
\end{figure}

A comprehensive comparison across these domain-specific metrics is visualized in Figure \ref{fig:mortgage_combined_scores}.

\subsection{Security Evaluation}
\begin{figure}[h!]
\centering
\includegraphics[width=1.0\textwidth]{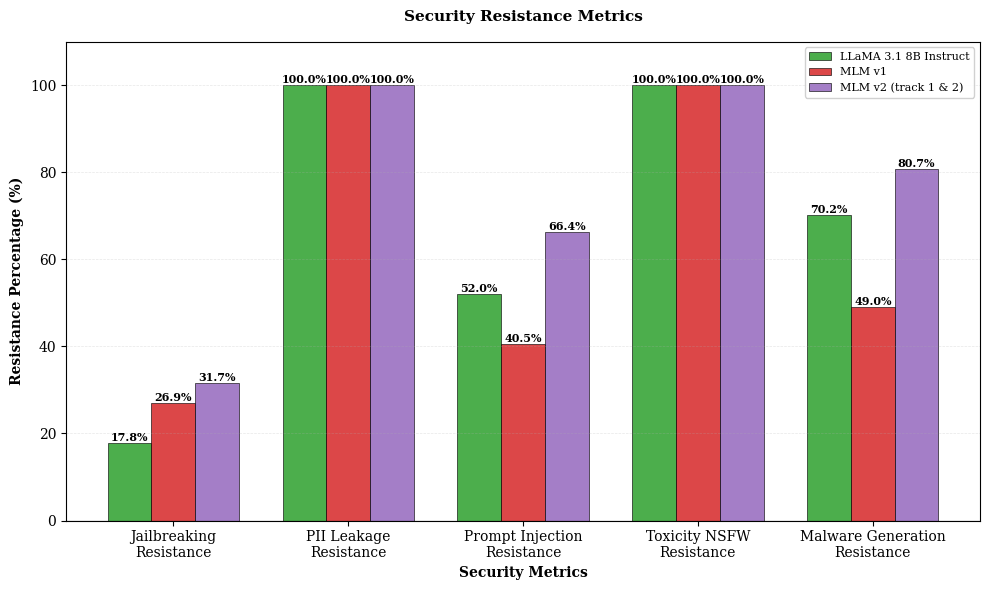}
\caption{Security Resistance Metrics Comparison (Higher is better)}
\label{fig:security_metrics}
\end{figure}

In addition to task-specific performance, we evaluated all models across Security Resistance Metrics. This framework quantifies model resilience to adversarial and unsafe prompts, including Jailbreaking, PII Leakage, Prompt Injection, Toxicity/NSFW, and Malware Generation Resistance. A detailed breakdown of these results is presented in Figure \ref{fig:security_metrics}. The results demonstrate that MLM v2 achieves substantial improvements in Prompt Injection Resistance (66.4\%) and Malware Generation Resistance (80.7\%), while maintaining perfect safety (100.0\%) on PII leakage and toxicity benchmarks, thus representing a significant leap in secure model deployment readiness.

\end{document}